\pdfoutput=1

\documentclass[11pt]{article}

\usepackage{acl}
\usepackage{hyperref}
\usepackage{times}
\usepackage{latexsym}
\usepackage{multirow}
\usepackage{booktabs}
\usepackage{microtype}
\usepackage{multirow}
\usepackage{latexsym}
\usepackage{booktabs}
\usepackage{courier}
\usepackage{amsmath}
\DeclareMathOperator*{\argmax}{argmax}
\usepackage{dsfont}
\usepackage{booktabs} 
\usepackage{siunitx} 
\usepackage{subfig}
\usepackage{placeins}
\usepackage{tikz}
\usepackage{hyperref}

\usepackage[T1]{fontenc}

\usepackage[utf8]{inputenc}

\usepackage{microtype}

\usepackage{inconsolata}

\title{SEME at SemEval-2024 Task 2: Comparing Masked and Generative Language Models on Natural Language Inference for Clinical Trials}

 \author{Mathilde Aguiar, Pierre Zweigenbaum, Nona Naderi \\
         Université Paris-Saclay, CNRS, Laboratoire Interdisciplinaire des Sciences du Numérique, 91400, Orsay, France  \\ 
         \texttt{\{mathilde.aguiar, pierre.zweigenbaum, nona.naderi\}@lisn.fr}}

\begin{document}
\maketitle
\begin{abstract}

This paper describes our submission to Task 2 of SemEval-2024: Safe Biomedical Natural Language Inference for Clinical Trials. The Multi-evidence Natural Language Inference for Clinical Trial Data (NLI4CT) consists of a Textual Entailment (TE) task focused on the evaluation of the consistency and faithfulness of Natural Language Inference (NLI) models applied to Clinical Trial Reports (CTR). We test 2 distinct approaches, one based on finetuning and ensembling Masked Language Models and the other based on prompting Large Language Models using templates, in particular, using Chain-Of-Thought and Contrastive Chain-Of-Thought. 
Prompting Flan-T5-large in a 2-shot setting leads to our best system that achieves 0.57 F1 score, 0.64 Faithfulness, and 0.56 Consistency.

\end{abstract}

\section{Introduction}

The digitization of medical documents allows the development of tools using various NLP techniques. In the case of Clinical Trial Reports (CTR), these tools can facilitate recruiting patients to participate in a trial or help researchers keep up to date with the literature. Natural Language Inference (NLI) is particularly useful in detecting the relationship between a CTR and a statement. For instance, it can be used for patient-trial matching.

Task~2 of \mbox{SemEval} 2024 defines a Textual Entailment (TE) task applied to English breast cancer CTRs. A submitted system must perform a binary classification based on a CTR and a given statement, using the labels \textit{entailment} or \textit{contradiction}. In addition to the traditional F1-measure for Textual Entailment, the submitted systems are evaluated on 2 strong metrics: Faithfulness and Consistency.

In this paper, we first introduce the task and some related work in Sec.~\ref{sec:Background}. Sec.~\ref{sec:system} describes our proposed approaches, while Sec.~\ref{sec:expe} gives further details about the experimental setup. Sec.~\ref{sec:results} presents the results and comparative analysis of methods, and Sec.~\ref{sec:conclusion} sums up our work done and provides ideas for future work.

\section{Background}
\label{sec:Background}

\subsection{Corpus and task description}

The NLI4CT \cite{jullien-etal-2024-semeval} corpus consists of a collection of breast cancer Clinical Trial Reports (CTR) taken from \href{https://clinicaltrials.gov/}{clinicaltrials.gov}. The documents are exclusively written in English.
These CTRs are structured with the following sections:
\textit{Intervention} section describes what treatment is going to be applied during the trial. \textit{Eligibility} section consists of a set of inclusion and exclusion criteria that a test subject must comply with. \textit{Results} section displays the outcome measures. Finally, \textit{Adverse Events} section describes the side effects and symptoms observed during the trial. In NLI4CT there are two types of instances: \textit{single}, where only 1 CTR is involved to perform the inference, and \textit{comparison} where 2 CTRs need to be compared.

The task's objective is to perform Natural Language Inference on these clinical trials.
A premise consists of a section of a CTR (or two CTRs if it is a comparison), and a statement is a single sentence. The model should predict whether the premise entails or contradicts the statement.
To tackle the NLI4CT task, the model must perform several kinds of inference, such as quantitative, common-sense, and medical reasoning (see Fig.~\ref{fig:nli_workflow}). The inference relationship can be predicted using the evidence, sentences where clues are contained, that are in one of the sections of a CTR. Evidence is provided only in the development and training sets. The dataset is balanced with half of the instances labeled as \textit{entailment} and the other half as \textit{contradiction} in the train and development subsets.

\subsection{Related work}
A previous edition of the NLI4CT task was run as SemEval 2023 Task~7 \cite{jullien-etal-2023-nli4ct}.  It was composed of 2 subtasks: an NLI classification task and an information retrieval task of evidence selection to support the predicted label. The training and development sets were the same as the present edition. For the first subtask, the task overview paper \cite{jullien-etal-2023-semeval} reports both generative and discriminative approaches for the submitted systems. Over the past few years, we have seen the fast-paced development of Large Language Models (LLMs) and their increased capabilities in addressing both generative and discriminative tasks. Even general-domain LLMs like Flan-T5-xxl in \citet{kanakarajan-sankarasubbu-2023-saama} and GPT-3.5 in \citet{pahwa-pahwa-2023-bphigh} have been achieving competitive performance on domain-specific tasks for the 2023 edition of the NLI4CT task.

\section{System overview}
\label{sec:system}
To address the NLI4CT task, we tested 2 main approaches: 
the first uses Pretrained Masked Language Models (MLM), and the second uses generative Large Language Models. We wanted to compare the ability of these two kinds of architectures to solve the same task, in particular in terms of consistency and faithfulness.

\subsection{Finetuning pretrained masked language models}

\begin{figure}[ht]
    \centering
    \includegraphics[width=\columnwidth]{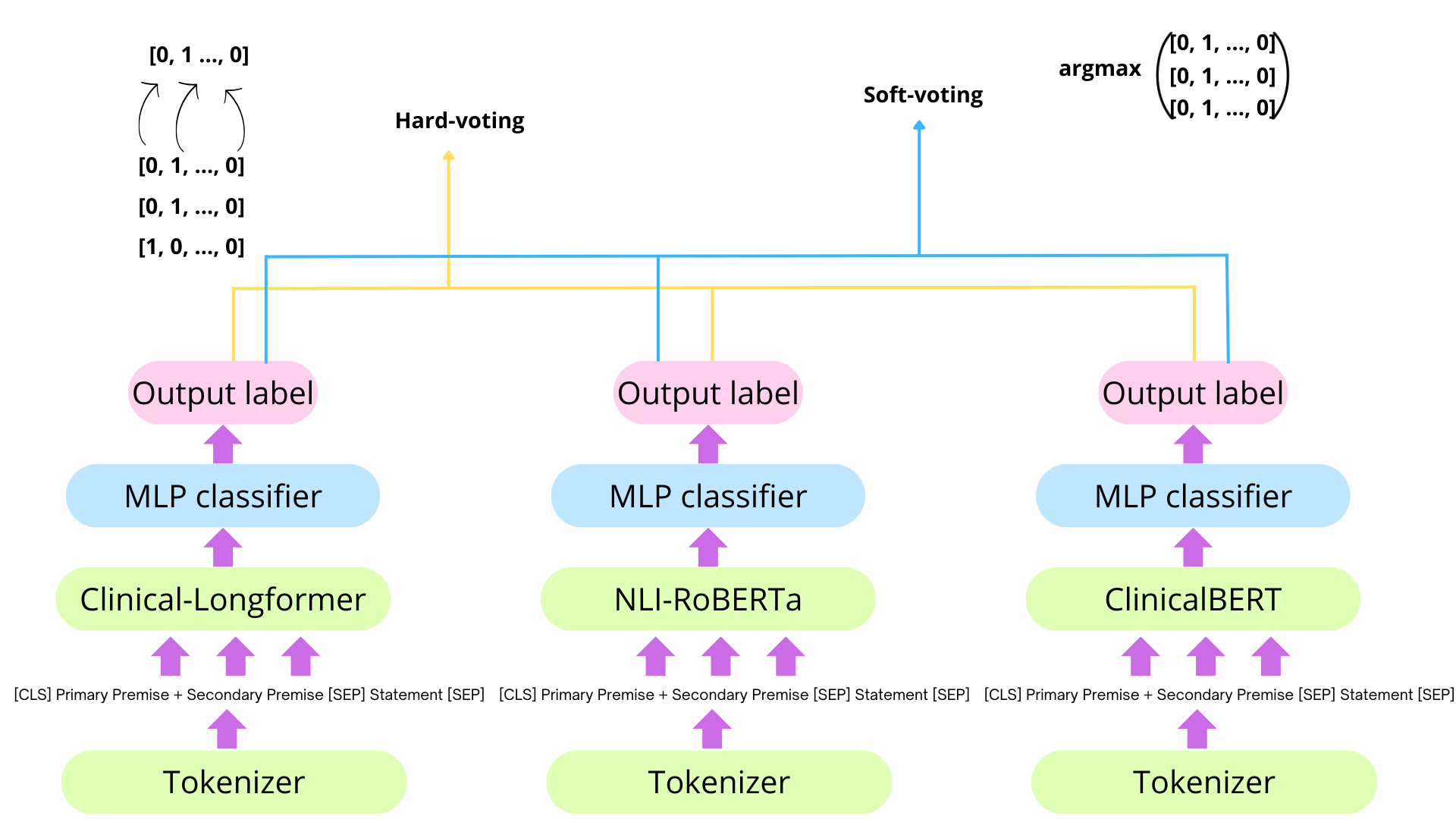}
    \caption{MLM ensemble architecture overview.}
    \label{fig:mlm_archi}
\end{figure}

Our first system is based on finetuning and ensembling multiple MLMs on the task data (see an example in Fig.~\ref{fig:mlm_archi}).
We first finetune each model using the train and development splits of NLI4CT. We evaluate each finetuned model on the test set. We perform experiments with two ensembling methods: hard-voting and soft-voting. The hard-voting method consists of selecting the label $y$ that gets the majority of votes across the predictions of each model $j$, defined as follows: 

$$ \Tilde{y} = \argmax_y \sum_{j=1}^{N} \mathds{1}(\Tilde{y}_j=y)  $$
Soft-voting is computed by using the argmax of probabilities $P_j$ from each model $j$ for a given label $y$: 
$$ \Tilde{y} = \argmax_y \sum_{j=1}^{N}P_j(y) $$

\subsection{Prompting generative large language models}
\label{sec:prompting_llm}
We designed a set of prompts that rely on the following techniques:

1. A simple prompt instructing the model to perform Textual Entailment, giving the statement and a premise composed of the whole section where the evidence comes from. We took inspiration from the instruction templates found in the Flan-Muffin dataset\footnote{\url{https://huggingface.co/datasets/causal-lm/flan-muffin}} that \cite{Lou2024MUFFIN} used to instruction-tune the Flan-T5 models \cite{chung2022scaling}. 
The template starts with optional demonstrations that instantiate this prompt with $n$ training or development examples in $n$-shot settings:

\emph{[\underline{Demonstrations}] [Premise] [Statement] Based on this premise, is the hypothesis true? OPTIONS: -'Yes' -'No'}

2. 
Using the concept of Chain-Of-Thought \citep{Wei2022ChainOT} that decomposes the reasoning behind a given example; we insert the premise sentences that are the actual evidence used to infer an entailment or a contradiction in the demonstrations. See ~\ref{sec:COT} for a detailed example.

3. 
We tested the related Contrastive Chain-Of-Thought (CCOT) \citep{chia2023contrastive} technique that gives both one correct and one incorrect explanation in addition to the original template. In our case, we inserted premise sentences that were not actual evidence. See ~\ref{sec:CCOT} for an example. CCOT is inspired by how humans learn from positive and negative examples and aims to reduce reasoning errors by indicating what mistakes to avoid.

For the demonstrations, we tried three few-shot settings: zero-shot (ZS:
no demonstration, only for the first template), 1-shot, and 2-shot.
See Appendix~\ref{sec:appendix_prompts} for detailed examples of the prompts.

\section{Experimental setup}
\label{sec:expe}

\subsection{Data pre-processing}

We used the NLI4CT train and development splits published by BigBio on HuggingFace\footnote{\url{https://huggingface.co/datasets/bigbio/sem_eval_2024_task_2}} and enriched them with new columns: primary and secondary evidence and premises from the JSON files provided by the organizers. We used this dataset to build our prompts (see Sec.~\ref{sec:prompting_llm}). We shuffled the train and dev sets and selected random instances to include as demonstrations in our 1 and 2-shot settings. 

\subsubsection{Ensembling MLMs}

We used Masked Language Models that are pretrained on general domain data or clinical data. For the general domain, we selected NLI-RoBERTa\footnote{\url{https://huggingface.co/sentence-transformers/nli-roberta-base-v2}} \cite{reimers-gurevych-2019-sentence} from Sentence Transformers, which has been previously finetuned for NLI using SICK \cite{marelli-etal-2014-sick} and STS benchmark \cite{cer-etal-2017-semeval}. For the clinical pretrained models we use Clinical-Longformer\footnote{\url{https://huggingface.co/yikuan8/Clinical-Longformer}} \cite{li2023comparative}, which can handle a context window up to 4096 tokens, and ClinicalBERT\footnote{\url{https://huggingface.co/medicalai/ClinicalBERT}} \cite{Wang2023clinicalbert} which has been pretrained on Electronic Health Records. We used Optuna \cite{optuna_2019} for hyperparameter search and set our final configuration with a learning rate of $5e^{-5}$ using the AdamW \cite{Loshchilov2017DecoupledWD} optimizer, a batch size of 64 and finetuned the models for 4 epochs. Ensembles of the same model used a different random seed when training each instance. We used 4 NVIDIA Tesla V100 with 32 GB of RAM with a training and inference time varying from 3 to 6.5~hours. A more detailed analysis of the training cost can be found in Appendix~\ref{sec:energy_conso}.

\subsubsection{Prompting generative LLMs}
We tested several Large Language Models (see Appendix~\ref{sec:appendix_compl_expe}).
We eventually chose Flan-T5-large\footnote{\url{https://huggingface.co/google/flan-t5-large}} \cite{chung2022scaling}
for its ability to output answers that are easier to parse than the longer and more challenging answers that could be provided by Llama-2 \cite{touvron2023llama} or Mistral \cite{jiang2023mistral}. 
Flan-T5 has been pretrained on a mixture of 473 datasets covering 1,836 tasks. However, it has no biomedical or clinical pretraining. We rely on the HuggingFace framework for all experiments. 
We used the same computing setup as in the previous set of experiments. The codebase for all of our experiments is freely available.\footnote{\url{https://github.com/MathildeAguiar/SemEval-2024-Task-2}}

\subsection{Evaluation}
We evaluate our models using the following metrics.
The F1 score of the \emph{Entailment} class is measured on a control set of the gold test set which is the same as the NLI4CT 2023's test data.
Faithfulness measures whether a model changes predictions when an `entailing' statement is changed into a `contradicting' statement.
Consistency measures whether a model keeps its predictions when a statement is changed while preserving its relation to the premise. Both metrics are computed on a contrast set of the gold test set that has undergone perturbations
(more details in \citet{jullien-etal-2024-semeval}).

\section{Results}
\label{sec:results}

\subsection{Quantitative analysis}

Under the username \textit{math\_agr}, our team ranked 27th for an F1 score of 0.57, 18th for Faithfulness of 0.64, and 25th for Consistency of 0.56. Tables~\ref{tab:all_res_mlm}--\ref{tab:error_rate_inf} report the results of our experiments on the test set.  

\begin{table}[h]
    \resizebox{\columnwidth}{!}
    {
    \setlength{\tabcolsep}{3pt}
        \begin{tabular}{lccc}
        \toprule
        Single system &   F1 & Faithfulness & Consistency\\
        \midrule
        Majority class & 0.67 & 0.00 & 0.38 \\
        tf.idf \cite{jullien-etal-2024-semeval} & 0.41 & 0.47 & 0.47 \\
        \textit{FZI-WIM} & 0.80 & 0.90 & 0.73 \\
        \textit{rezazzr} & 0.06 & 0.95 & 0.60 \\
        \textit{NYCU-NLP} & 0.78 & 0.92 & 0.81 \\
        \midrule
        a: NLI-RoBERTa  & 0.56  & 0.58 & \underline{0.57} \\
        b: ClinicalBERT & 0.00 & \textbf{1.00} & \textbf{0.62}\\
        c: Clinical-Longformer & \textbf{0.67} & 0.00 & 0.38\\
        \end{tabular}
}
    \resizebox{\columnwidth}{!}
    {
        \begin{tabular}{lccc}
        \toprule
        Ensemble
        & s/h & s/h & s/h \\
        \midrule
        (a+a+a)  &0.57/0.57 & 0.58/0.54& \underline{0.57}/0.56\\
        (b+b+b)  & 0.56/0.63 & 0.37/0.16 & 0.47/0.43\\
        (c+c+c)  & \textbf{0.67}/0.64& 0.00/0.09 & 0.38/0.40\\      
        d: (a+b+c)  & 0.55/0.57 & 0.45/0.40 & 0.52/0.52 \\
        \midrule
        (d) + Flan-T5-large  & 0.57 (h) &\underline{0.64} (h) & 0.56 (h) \\
        \bottomrule
        \end{tabular}
   }
    \caption{F1 score, Faithfulness, and Consistency for single Masked Language Models then soft (s) and hard (h) ensembling. Ensembles such as (a+a+a) consist of 3 instances of the same model. Flan-T5-large is used in a 2S setting (see Tab.~\ref{tab:all_res_llm} below).}
    \label{tab:all_res_mlm}
\end{table}

Each model has different strengths and weaknesses across the three metrics
in the MLM experiments.
The single NLI-RoBERTa seems to be the most stable baseline despite its lack of pretraining on biomedical data. It has already been finetuned on general-domain NLI, and its sentence-level representation seems to boost its performance.
The ensemble of 3 NLI-RoBERTa does not add enough diversity to improve its results.
The single ClinicalBERT obtains an F1-score of 0.00: we observed that it always predicts the label \emph{Contradiction}, which causes a precision and recall of 0.00. Faithfulness yields 1.00 because it is computed on instances of the contrast test set that are all labeled as \emph{Contradiction}.  
The ensemble of 3 ClinicalBERT does not have this issue: some seeds led to better models.
The single Clinical-Longformer obtains the best results in terms of F1-score but the worst on the other two metrics, especially on Faithfulness. It predicts almost exclusively \emph{Entailment}, which leads to Faithfulness and Consistency complementary to ClinicalBERT's. 
The ensemble keeps the same issues.
An ensemble (d) of the three single models could not improve the single NLI-RoBERTa.
Adding Flan-T5's 2-shot predictions to the ensemble increased Faithfulness by 0.24 points but did not yield better F1. This did not improve either over Flan-T5 alone (see row 2S in Tab.~\ref{tab:all_res_llm}).

\begin{table}[h]
    {\centering
        \begin{tabular}{lccc}
        \toprule
        Prompt &  F1  & Faithfulness & Consistency\\
        \midrule
        ZS  &  \underline{0.56} & 0.57 & 0.55 \\
        1S & 0.53 & 0.63 & \textbf{0.57} \\
        2S  & \textbf{0.57} & 0.64 & 0.56 \\
        \midrule
        1SCOT  & 0.39 & 0.70 & 0.53 \\
        2SCOT  & 0.43 & 0.69 & 0.51 \\
        \midrule        
        1SCCOT  & 0.28 & \textbf{0.85} & \textbf{0.57}\\
        2SCCOT  & 0.24 & \underline{0.81} & 0.56 \\
        \bottomrule
        \end{tabular}\par
    }
    \caption{F1 score, Faithfulness, and Consistency for the LLM approach, using Flan-T5-large.}
    \label{tab:all_res_llm}
\end{table}

Prompting Flan-T5-large in few-shot mode performs as well as the fine-tuned NLI-RoBERTa.  Increasing the number of demonstrations tends to improve the scores.
This illustrates the usual trade-off between fine-tuning a smaller model or prompting a larger model without fine-tuning it.
The Chain-Of-Thought method makes it more difficult to recognize \emph{Entailment} relations and leads to lower F1.  As seen above, this mechanically increases Faithfulness.
Contrastive Chain-Of-Thought further reduces the number of predicted \emph{Entailment} relations, with an associated increase in Faithfulness.
All systems achieve similar Consistency. 
The tf-idf baseline was provided by the task organizers.  Some of our proposed systems scored below the baseline in some metrics.  For instance, Clinical-Longformer obtained a much lower Faithfulness and Consistency, 
ClinicalBERT, CCOT, and 1SCOT prompts obtained lower F1 scores.
According to the leaderboard, the top scores were
0.80 for the F1 score,
0.95 for Faithfulness, 
and 0.81 for Consistency,  
achieved by 3 different teams.
We do not have information regarding the approaches these teams chose at the time of writing.
Using last year's results on the F1 score, the approach of \citet{kanakarajan-sankarasubbu-2023-saama}, using Flan-T5-xxl, achieved an F1 score of 0.83. Their approach differs from ours by not only prompting Flan-T5 but by finetuning it beforehand using single- and multiple-instruction templates. This approach leads to a boost in performance compared to our simpler approach. \citet{takehana-etal-2023-stanford} also performed ensembling and voting of MLMs and achieved an F1 score of 0.66. They performed what we called `hard voting,' using 10 models for their ensemble and performing data augmentation on the original task dataset. Their result is comparable to our approach using an ensemble of 3 ClinicalBERT or 3 Clinical-Longformer.

\subsection{Error analysis}
\newcommand{\systems}{Systems: ensemble of 3 NLI-RoBERTa; ensemble of all MLM baselines (d) + Flan-T5-large (2S); Flan-T5-large in 2-shot (2S) and 1-shot contrastive chain-of-thought (1SCCOT) settings.}
\newcommand{\systseeabove}{Systems: see Tab.~\ref{tab:error_rate_labels}.}

In this section, we analyze our models in more depth by breaking down their results according to
gold labels,
whether a comparison of CTRs is involved,
the types of inference to perform,
CTR sections,
and examine the F1 score per intervention type.
For simplicity, we focus our analysis only on the two best-performing systems of each approach.    

\paragraph{Accuracy per gold label}
From the accuracy displayed in Tab.~\ref{tab:error_rate_labels}, we observe that our LLM methods, especially CCOT, handle the Contradiction examples better. 
This label is the most frequent in the test set (67\% of instances labeled as Contradiction and 33\% as Entailment). MLMs, in contrast, have similar accuracy across both labels.
\begin{table}[h]
    {\centering
        \begin{tabular}{lcc}
        \toprule
        System & Entailment & Contradict. \\
        \midrule
        3 NLI-RoBERTa  & \textbf{55} & 56  \\
        (d) + Flan-T5-large & \textbf{55} & 48 \\
        \midrule
        2S  & 44 &  64 \\
        1SCCOT & 20 & \textbf{82} \\
        \bottomrule
        \end{tabular}\par
   }
    \caption{Accuracy (in \%) per label: \emph{Entailment} and \emph{Contradiction} (Contradict.). \systems}
    \label{tab:error_rate_labels}
\end{table}

\paragraph{Comparison versus Single}
The \emph{Comparison} of 2 CTRs implies longer input sequences and possibly an increased complexity since the model needs to confront the elements of two separate documents. Surprisingly, as reported in Tab.~\ref{tab:error_rate_compa}, we observe that all models perform similarly for \emph{Comparison} and \emph{Single}. We can hypothesize that the models are able to find more clues with 2 documents instead of 1 and predict more accurate labels.

\begin{table}[h]
    {\centering
    \setlength{\tabcolsep}{3pt}
        \begin{tabular}{lcc}
        \toprule
        System & Single & Comparison \\
        \midrule
        3 NLI-RoBERTa  &  56 &  56 \\
        (d) + Flan-T5-large & 49 & 51 \\
        \midrule
        2S  & 59 &  56 \\
        1SCCOT  & \textbf{61} & \textbf{61} \\
        \bottomrule
        \end{tabular}\par
   }
    \caption{Accuracy (in \%) per CTR type: \emph{Single} and \emph{Comparison}. \systseeabove}
    \label{tab:error_rate_compa}
\end{table}

\paragraph{CTR sections}
From the accuracy displayed in Tab.~\ref{tab:error_rate_section}, we observe no performance distinction between the models for different sections. 

\begin{table}[h]
    {\centering
    \setlength{\tabcolsep}{3pt}
        \begin{tabular}{lcccc}
        \toprule
        System & AE & Int. & Elig. & Res. \\
        \midrule
        3 NLI-RoBERTa  & 60 & 59 & 52 & 52 \\
        (d) + Flan-T5-large & 43 & 46 & 55 & 57\\
        \midrule
        2S  & 55 & 58 & \textbf{61} & 54 \\
        1SCCOT  & \textbf{62} & \textbf{63} & 58 & \textbf{60} \\
        \bottomrule
        \end{tabular}\par
   }
    \caption{Accuracy (in \%) per CTR section: \emph{Adverse events} (AE), \emph{Intervention} (Int.), \emph{Eligibility} (Elig.), and \emph{Results} (Res.). \systseeabove}
    \label{tab:error_rate_section}
\end{table}

\paragraph{Types of `intervention'}
Tab.~\ref{tab:error_rate_inf} results were obtained directly from the task organizers' evaluation script. 
Once again NLI-RoBERTa is stable across \emph{Paraphrase} and \emph{Definition} interventions and achieves the best performance. NLI-RoBERTa seems to be less sensitive to semantic change when it comes to paraphrasing.
Its score for \emph{Definition} shows that it can capture the relevant information better when more details are provided. Contrastive Chain-Of-Thought does not increase the model's resistance to semantic change (as shown by the results on \emph{Paraphrase}), its ability to perform numerical inference (see results on Numerical paraphrase) or to focus on relevant information (see results on \emph{Definition}). For the latter, the model might struggle to focus on relevant information because of the long length of the input prompts (see Tab.~\ref{table:metrics_input_llm}). 

\begin{table}[h]
    {\centering
    \setlength{\tabcolsep}{3pt}
        \begin{tabular}{lccccc}
        \toprule
        System & Def. & NP & Para.  \\
        \midrule
        3 NLI-RoBERTa  &  \textbf{0.57} & \textbf{0.51} & \textbf{0.56}\\
        (d) + Flan-T5-large & 0.39 & 0.46 & 0.54\\
        \midrule
        2S  & 0.39 & 0.46 & 0.54  \\
        1SCCOT  & 0.31 & 0.26 & 0.25 \\
        \bottomrule
        \end{tabular}
    
   }
    \caption{F1 score per intervention type: \emph{Definition} (Def.), \emph{Numerical Paraphrase} (NP), or \emph{Paraphrase} (Para.) interventions. \systseeabove}
    \label{tab:error_rate_inf}
\end{table}

\section{Conclusion and future work}
\label{sec:conclusion}

This paper describes the two systems proposed by the SEME team for the SemEval 2024 Task~2 NLI4CT. Our first approach is based on the finetuning and ensembling of Masked Language Models, using only the challenge's data. Our second approach consists of a pipeline to prompt Large Language Models, using prompt engineering techniques, such as Chain-Of-Thought and Contrastive Chain-of-Thought, in Zero-shot, 1-shot, and 2-shot manners. Our two best-reported results are 0.57 F1 score, 0.64 Faithfulness, and 0.56 Consistency, with prompting Flan-T5-large in a 2-shot manner, ranking 27th out of 32 submissions for F1, 18th for Faithfulness and 25th for Consistency. We obtain the same scores for the MLM system using an ensemble composed of a finetuned NLI-RoBERTa + Clinical-Longformer + ClinicalBERT + the predictions of Flan-T5-large, that is 0.57 for F1 score, 0.64 for Faithfulness, and 0.56 for Consistency.

Some future work could include the continuation of the Masked Language Models pretraining on unlabeled clinical trials, before performing a similar finetuning as presented in the paper. We could also apply this approach to medical Large Language Models like MEDITRON \cite{chen2023meditron70b}, by performing instruction-tuning using clinically oriented instructions and then prompting the resulting model on the task data. Another possible approach, similar to \cite{conceicao-etal-2023-lasigebiotm}, would be to incorporate domain ontologies (like UMLS) into the finetuning of Masked Language Models to provide definitions and supplementary knowledge.

\section*{Ethical statement}
The NLI4CT task uses clinical data extracted and processed from \url{https://clinicaltrials.gov/}. This resource is freely available, provided by the National Library of Medicine, and is an official U.S. Department of Health and Human Services website.

\subsection*{Carbon emissions}
\label{sec:energy_conso}
Another arguable ethical aspect of our approach is the carbon emissions generated by our models' training and inference. Our experiments used 4 Tesla V100 GPUs paired with 2 Intel Xeon Gold 6148 20 cores and 384 GB of RAM. Depending on the approach chosen, the running time can be up to 10 times longer. For instance, we observe an execution time of 3 hours for the training and inference of an ensemble of 3 ClinicalBERT models. For the inference of Flan-T5-large on a 2-shot Contrastive Chain-Of-Thought, we achieve up to 30 hours of running time to get the predictions for all instances of the test set. Globally, we can say that the MLM approach is computationally more efficient, with running times varying from 3 to 6.5 hours (for the ensemble of ClinicalBERT, NLI-RoBERTa, and Clinical-Longformer). For the LLM approach, we observe running times ranging from 10.5 hours (in Zero-shot) to 38 hours (in 1-shot Chain-Of-Thought).

We used Green Algorithms\footnote{\url{http://calculator.green-algorithms.org/}} \cite{Lannelongue_Grealey_Inouye_2021} to estimate carbon emissions, 
taking into consideration our aforementioned computational configuration. The MLM approach produces up to 831g of $CO_2$ with the 3 models ensembling approach. For the LLM approach, the emissions vary from 1.34 kg of $CO_2$ for zero, 1, and 2-shot experiments to 4.86kg for Contrastive Chain-Of-Thought experiments.

Considering the little gain in performance of LLMs compared to MLMs using our approach and the $CO_2$ overconsumption of the LLMs, it would be more reasonable to use the MLM approach in our case. The MLM approach also provides faster predictions, which can be much more convenient. 

\section*{Acknowledgements}
This work benefited from the GPUs provided by Lab-IA, an institution member of Université Paris-Saclay. This work was also supported through the CNRS grant 80|PRIME.

\bibliography{acl_latex}

\appendix

\section{Hyperparameters}
Tab.~\ref{tab:hyperparams_mlm} shows the final hyperparameters used for finetuning the Masked Language Model systems.
\begin{table}[h]
    {\centering
        \begin{tabular}{lccc}
        \toprule
        Hyperparameter & Value \\
        \midrule
        Nb. epochs  &  4 \\
        Batch size & 64 \\
        Learning rate  & $5e-5$ \\
        Optimizer  &  AdamW\\
        \bottomrule
        \end{tabular}\par
    }
    \caption{Hyperparameters to finetune the MLM systems.}
    \label{tab:hyperparams_mlm}
\end{table}

\section{Example of Natural Language Inference mechanism}
\begin{figure*}[htb]
    \centering
    \includegraphics[width=.7\linewidth]{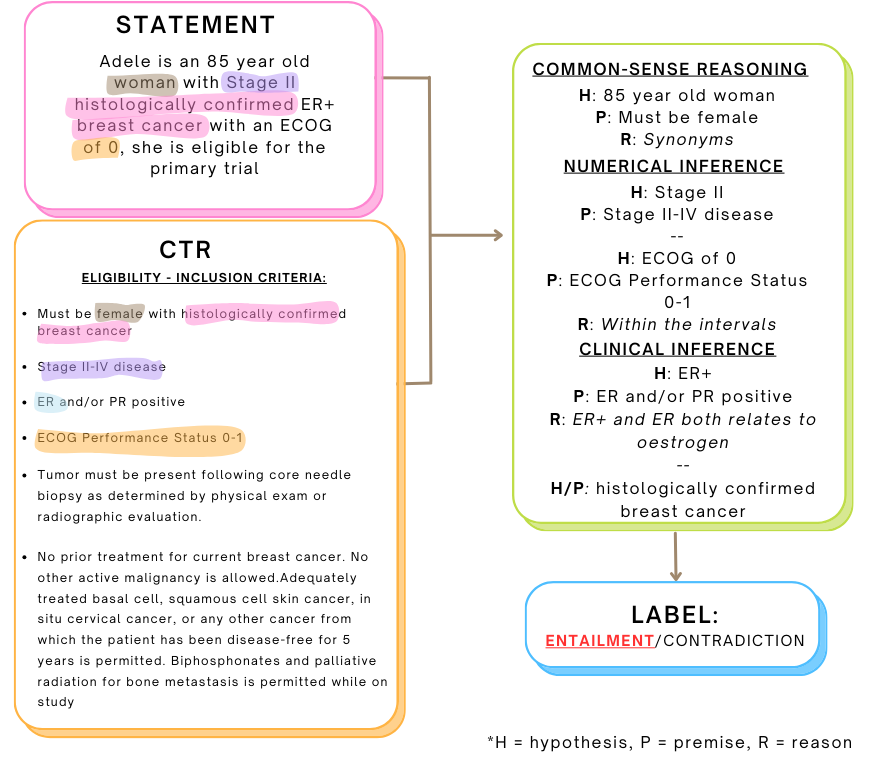}
    \caption{Example of an inference mechanism using a statement and the \emph{Eligibility} section of a CTR.}
    \label{fig:nli_workflow}
\end{figure*}

Fig.~\ref{fig:nli_workflow} shows an example of the kinds of inference performed by the NLI system in order to predict the correct label. 

\newpage
\section{Prompts}
\label{sec:appendix_prompts}

\subsection{Simple prompt}
\label{sec:simple_prompt}
\begin{figure}[h]
    \centering
    \includegraphics[width=\columnwidth]{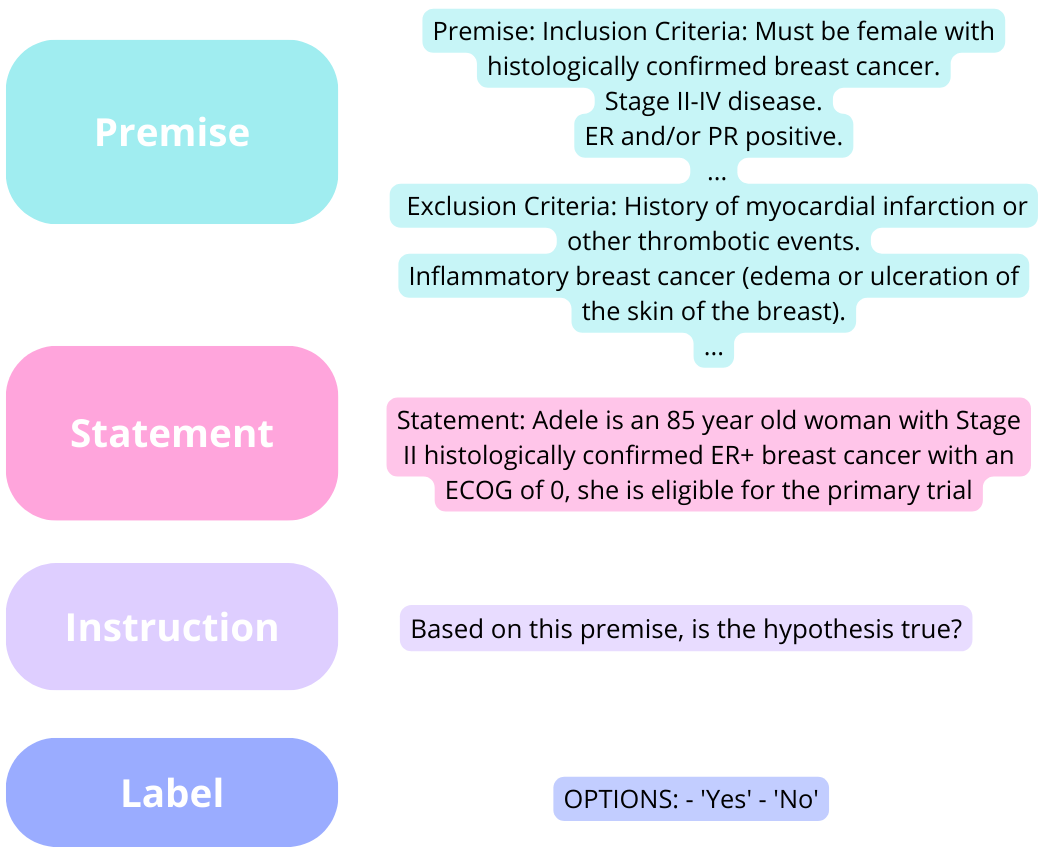}
    \caption{Example Zero-shot prompt.}
    \label{fig:ex_prompt}
\end{figure}

Fig.~\ref{fig:ex_prompt} displays an example Zero-shot prompt.
For $n$-shot prompts, we insert $n$ demonstrations before this prompt.  Each demonstration is built from training data; in a demonstration, the Label part is replaced with `Answer: Yes' or `Answer: No' depending on whether the example's label is \emph{Entailment} or \emph{Contradiction}.

\subsection{Chain-Of-Thought}
\label{sec:COT}

\begin{figure}[h]
    \centering
    \includegraphics[width=\columnwidth]{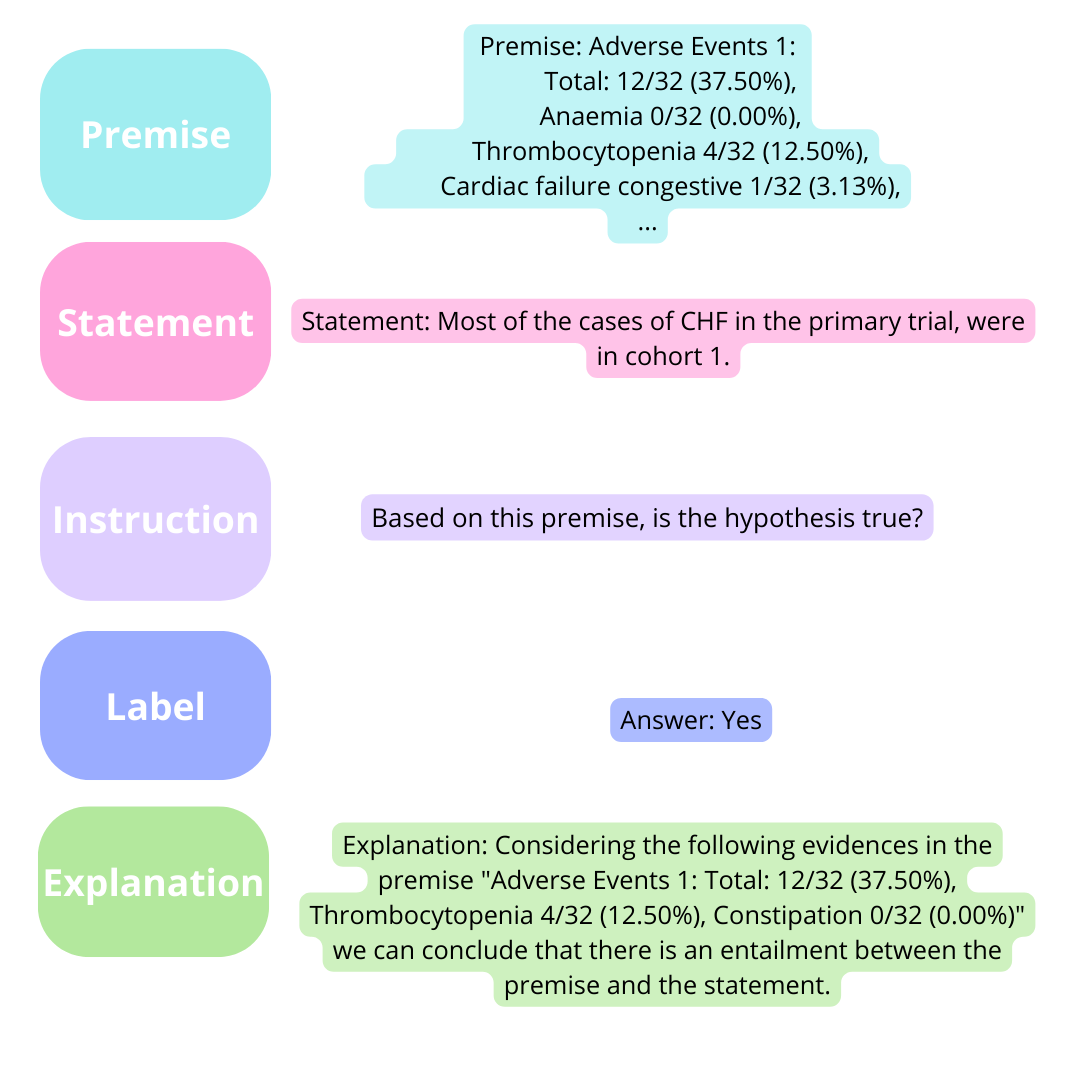}
    \caption{Example Chain-Of-Thought demonstration.
    }
    \label{fig:ex_cot}
\end{figure}

Fig.~\ref{fig:ex_cot} displays an example Chain-Of-Thought demonstration. Our initial demonstrations are modified to include the idea of Chain-Of-Thought as mentioned in \citet{Wei2022ChainOT}.

\subsection{Contrastive Chain-Of-Thought}
\label{sec:CCOT}

\begin{figure}[h]
    \centering
    \includegraphics[width=\columnwidth]{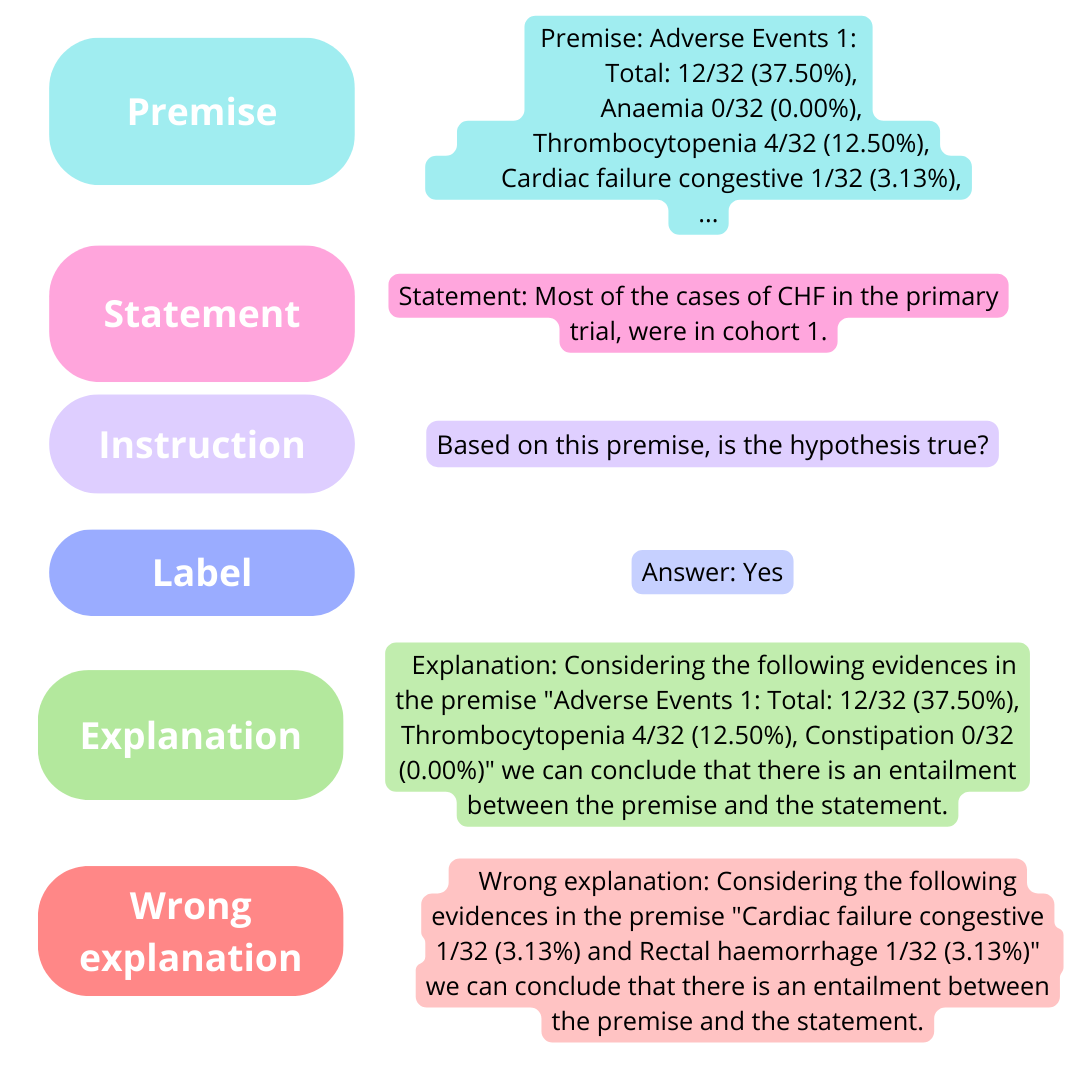}
    \caption{Example Contrastive Chain-Of-Thought demonstration.
    }
    \label{fig:ex_ccot}
\end{figure}

Fig.~\ref{fig:ex_ccot} displays an example of our Contrastive Chain-Of-Thought prompt. Our initial demonstrations are modified to include the idea of a Contrastive Chain-Of-Thought as mentioned in \citet{chia2023contrastive}.

\section{NLI4CT dataset statistics}
Tab.~\ref{tab:dataset_stats} shows statistics regarding the original task's data, such as the number of CTRs, of statements, the average length of a statement or evidence, and the max length of an evidence or statement.

\begin{table}[h]
\centering
    {
        \begin{tabular}{lc}
        \toprule
        Metric & Value  \\
        \midrule
        Nb. CTRs (documents)  &  999  \\
        Nb. statements & 2,400   \\
        Avg. length statement & 19.5 \\
        Max. length statement & 65 \\
        Avg. length evidence & 10.7 \\
        Max. length evidence & 197 \\
        \bottomrule
        \end{tabular}
   }
    \caption{Statistics about the NLI4CT train and dev sets.}
    \label{tab:dataset_stats}
\end{table}

\begin{table}[h]
\centering
    {
        \begin{tabular}{lcc}
        \toprule
        Subset & \emph{Entailment} & \emph{Contradiction}  \\
        \midrule
        Train  &  850 & 850  \\
        Validation & 100 & 100  \\
        Gold test set (whole) & 1841  & 3659 \\
        Gold test set (control set) & 250 & 250 \\
        Gold test set (contrast set) & 1591 & 3409\\
        \bottomrule
        \end{tabular}
   }
    \caption{Statistics about the number of \emph{Contradiction} and \emph{Entailment} instances in NLI4CT dataset.}
    \label{tab:dataset_stats2}
\end{table}

\section{Metrics on input sequences}
\subsection{MLM system input sequences}
Tab.~\ref{tab:metrics_input_mlm} displays the average, maximum, and minimum length of input sequences for the finetuning of MLMs.

\begin{table}[t]
\centering
{
\begin{tabular}{@{}lr@{}}
\toprule
\textbf{Metric} & \textbf{Value} \\
\midrule
     Mean nb. tokens &  480     \\
     Min. nb. tokens &   41    \\
     Max. nb. tokens & 2799   \\
\bottomrule
\end{tabular}%
}
\caption{Average, minimum, and maximum number of tokens of an input sequence for the MLM approach.}
\label{tab:metrics_input_mlm}
\end{table}

\subsection{LLM system input sequences}
Tab.~\ref{table:metrics_input_llm} displays the average, maximum, and minimum length of prompts used in Flan-T5.

\begin{table}[t]
\centering
{
\begin{tabular}{@{}lrrr@{}}
\toprule
\textbf{Prompt} & \textbf{Mean} & \textbf{Min.} & \textbf{Max.} \\
\midrule
     ZS &  573   & 92 & 1367  \\
     1S &   1650 & 835 & 3009   \\
     2S & 3036 & 1397& 6669  \\
     1S COT & 2474 & 1300& 6611 \\
     2S COT &  3933   & 6354 & 2484 \\
     1S CCOT & 2622  & 4285 & 1613   \\
     2S CCOT & 4826  & 3153 & 8321   \\
\bottomrule
\end{tabular}%
}
\caption{Average, minimum, and maximum numbers of tokens of each kind of prompt for the LLM approach.}
\label{table:metrics_input_llm}
\end{table}

\section{Prompt selection}

Tab.~\ref{tab:templates} displays the templates tried in order to find the one that would perform the best. The last two prompts were tested using Llama-2 and Mistral. The last prompt uses the concept of `persona prompting' \cite{zhang-etal-2018-personalizing} where we assign the LLM a role. 

\label{sec:appendix_compl_expe}

\begin{table*}[t]
\centering
\begin{tabular}{p{1cm} p{15cm}}
\toprule
\textbf{Id} & \textbf{Template}  \\
\midrule
1 & [Premise] [Statement] Does the premise entail the hypothesis? [Options] \\
\midrule
2 & [Premise] [Statement] Is the hypothesis entailed by the premise? [Options]  \\
\midrule
3 & [Premise] [Statement] If this premise is true, what does that tell us about whether it entails the hypothesis? [Options]  \\
\midrule
4 & From the following statement and premise, would you say there is a contradiction or an entailment between the statement and the premise? Just answer by saying 'contradiction' or 'entailment'. [Statement] [Premise]  \\
\midrule
5 & Imagine you are a medical practitioner and you are reviewing clinical trials. You are given a statement and a premise. You should determine if there is an entailment or a contradiction between the premise and the statement. There is necessarily an entailment or a contradiction, no neutral case. From the following statement and premise, would you say there is a contradiction or an entailment between the statement and the premise? Just answer by saying 'contradiction' or 'entailment'. [Statement] [Premise]   \\

\bottomrule
\end{tabular}
\caption{Other prompts tested on the LLM baselines.}
\label{tab:templates}
\end{table*}

\end{document}